%% file: main_ral_final.tex
\documentclass[letterpaper, 10 pt, journal, twoside]{IEEEtran}  

\IEEEoverridecommandlockouts
\usepackage{graphics} % for pdf, bitmapped graphics files
\usepackage{amsmath} % assumes amsmath package installed
\usepackage{amssymb}  % assumes amsmath package installed
\usepackage{cite}
\usepackage{amsfonts}
\usepackage{bm}
\usepackage{breqn}
\renewcommand{\vec}{\mathbf} % use this for non-italic non-greek vectors

\usepackage{multirow}
\usepackage{diagbox}
 \usepackage[subtle,tracking=normal]{savetrees}

\usepackage{balance}
\usepackage{caption}

\usepackage{hyperref}
\usepackage[percent]{overpic}
\usepackage{xcolor}

\newcommand\imgHalf{0.35}
\newcommand\imgHalfSpace{0.05}

\usepackage{multicol}
\usepackage{subcaption}
\usepackage{graphicx}

\usepackage{soul}

\usepackage{hyperref}
\usepackage[percent]{overpic}
\usepackage{xcolor}
\newcommand{\insertYoutubeLink}{\url{https://youtu.be/xqXHLzLsEV4}}

\title{
CPG-RL: Learning Central Pattern Generators\\ for Quadruped Locomotion
}
\author{Guillaume Bellegarda and Auke Ijspeert% 
\thanks{Manuscript received: May, 12, 2022; Revised August, 15, 2022; Accepted September, 12, 2022.}
\thanks{This paper was recommended for publication by Editor Dana Kulic upon evaluation of the Associate Editor and Reviewers' comments. 
This work was supported by the Swiss National Science Foundation
(SNSF) as part of project No.197237. } 
\thanks{Guillaume Bellegarda and Auke Ijspeert are with the BioRobotics Laboratory, Ecole Polytechnique Federale de Lausanne (EPFL).
        {\tt\footnotesize guillaume.bellegarda, auke.ijspeert@epfl.ch}}
\thanks{Digital Object Identifier (DOI) 10.1109/LRA.2022.3218167}
}

\begin{document}
\bstctlcite{MyBSTcontrol}
\maketitle

\markboth{IEEE Robotics and Automation Letters. Preprint Version. Accepted September, 2022}
{Bellegarda \MakeLowercase{\textit{et al.}}: CPG-RL: Learning Central Pattern Generators for Quadruped Locomotion}

%%%%%%%%%%%%%%%%%%%%%%%%%%%%%%%%%%%%%%%%%%%%%%%%%%%%%%%%%%%%%%%%%%%%%%%%%%%%%%%%
% Abstract
%%%%%%%%%%%%%%%%%%%%%%%%%%%%%%%%%%%%%%%%%%%%%%%%%%%%%%%%%%%%%%%%%%%%%%%%%%%%%%%%
\begin{abstract}
In this letter, we present a method for integrating central pattern generators (CPGs), i.e. systems of coupled oscillators, into the deep reinforcement learning (DRL) framework to produce robust and omnidirectional quadruped locomotion. The agent learns to directly modulate the intrinsic oscillator setpoints (amplitude and frequency) and coordinate rhythmic behavior among different oscillators. This approach also allows the use of DRL to explore questions related to neuroscience, namely the role of descending pathways, interoscillator couplings, and sensory feedback in gait generation. We train our policies in simulation and perform a sim-to-real transfer to the Unitree A1 quadruped, where we observe robust behavior to disturbances unseen during training, most notably to a dynamically added 13.75 kg load representing 115\% of the nominal quadruped mass. We test several different observation spaces based on proprioceptive sensing and show that our framework is deployable with no domain randomization and very little feedback, where along with the oscillator states, it is possible to provide only contact booleans in the observation space. Video results can be found at \insertYoutubeLink.
\end{abstract}


%%%%%%%%%%%%%%%%%%%%%%%%%%%%%%%%%%%%%%%%%%%%%%%%%%%%%%%%%%%%%%%%%%%%%%%%%%%%%%%%
% Introduction
%%%%%%%%%%%%%%%%%%%%%%%%%%%%%%%%%%%%%%%%%%%%%%%%%%%%%%%%%%%%%%%%%%%%%%%%%%%%%%%%
\section{Introduction}
\label{sec:introduction}

\IEEEPARstart{M}{uch} progress has been made in both understanding and replicating animal locomotion through robotics. Successful implementations include biologically-inspired controllers such as Central Pattern Generators (CPGs)~\cite{ijspeert2008,Fukuoka2003,owaki2013simple,owaki2017quadruped}, model-based control~\cite{dicarlo2018mpc,kim2019highly,sombolestan2021adaptive,bellicoso2018dynamic}, and learning-based approaches~\cite{tan2018minitaur,hwangbo2019anymal,kumar2021rma,lee2020anymal}. However, despite these successes and progress, the exact processes animals use to learn and produce motion remain unclear, especially related to how higher parts of the brain interact with CPGs in the spinal cord. The agility of robots also does not yet match that of animals. In this work, we draw inspiration from several of these areas to achieve legged locomotion by combining biology-inspired oscillators with the strength of deep learning. 

%--------------------------------------------- Related Work
\subsection{Related Work} 
%--------------------------------------------- Biology-Inspired Control
\subsubsection{Biology-Inspired Control} 
Central Pattern Generators are neural circuits located in the spinal cords of vertebrate animals that are capable of producing coordinated patterns of high-dimensional rhythmic output signals, with evidence from nature and experimentally validated on various robot hardware~\cite{ijspeert2008}. Among quadrupeds, the combination of CPGs, sensory input, reflexes, and mechanical design has led to adaptive dynamic walking on irregular terrain~\cite{Fukuoka2003}. Other works have focused on mechanical design inspired from biology for dynamic trot gaits~\cite{sprowitz2013cheetah}. Owaki et al.~show evidence that gait generation and transitions can occur through simple force feedback, without any explicit coupling between oscillators~\cite{owaki2013simple,owaki2017quadruped}. Righetti and Ijspeert added feedback from touch sensors to stop or accelerate transitions between swing/stance phases~\cite{righetti08}. The oscillators can also be formed in task space and mapped to joint commands with inverse kinematics, with additional control for push recovery and attitude control~\cite{barasuol2013reactive}. Ajallooeian et al.~augmented CPGs with both reflexes and Virtual Model Control~\cite{mos2013cat,mos2013oncilla}, and Hyun et al. presented a hierarchical controller leveraging proprioceptive impedance control for highly dynamic trot running~\cite{hyun2014trot}. Sensory feedback has not just been limited to proprioceptive information, for example Gay et al. used both a camera and gyroscope as feedback to a CPG to learn to walk over varying terrains~\cite{gay2013learncpg}.

\begin{figure}[!t]
      \centering
      \includegraphics[width=0.5\linewidth]{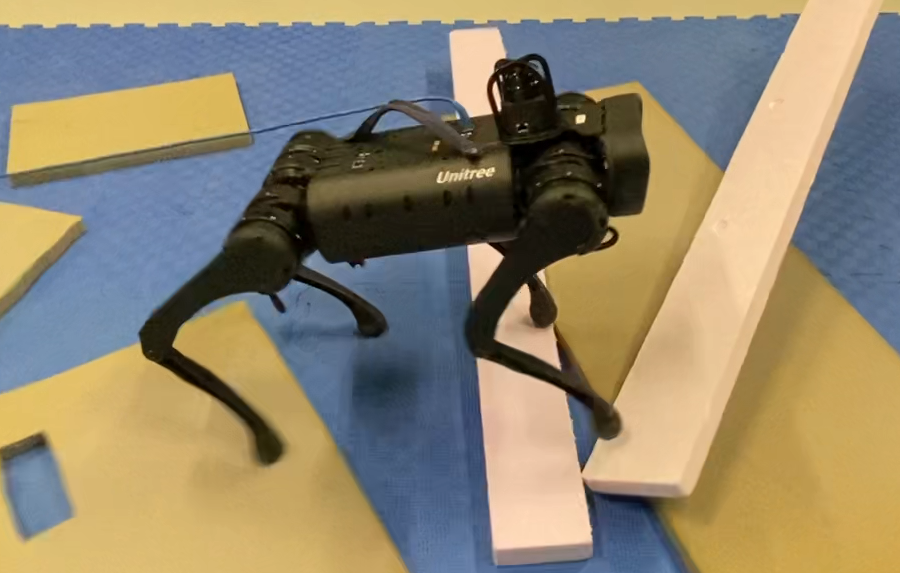}\includegraphics[width=0.5\linewidth]{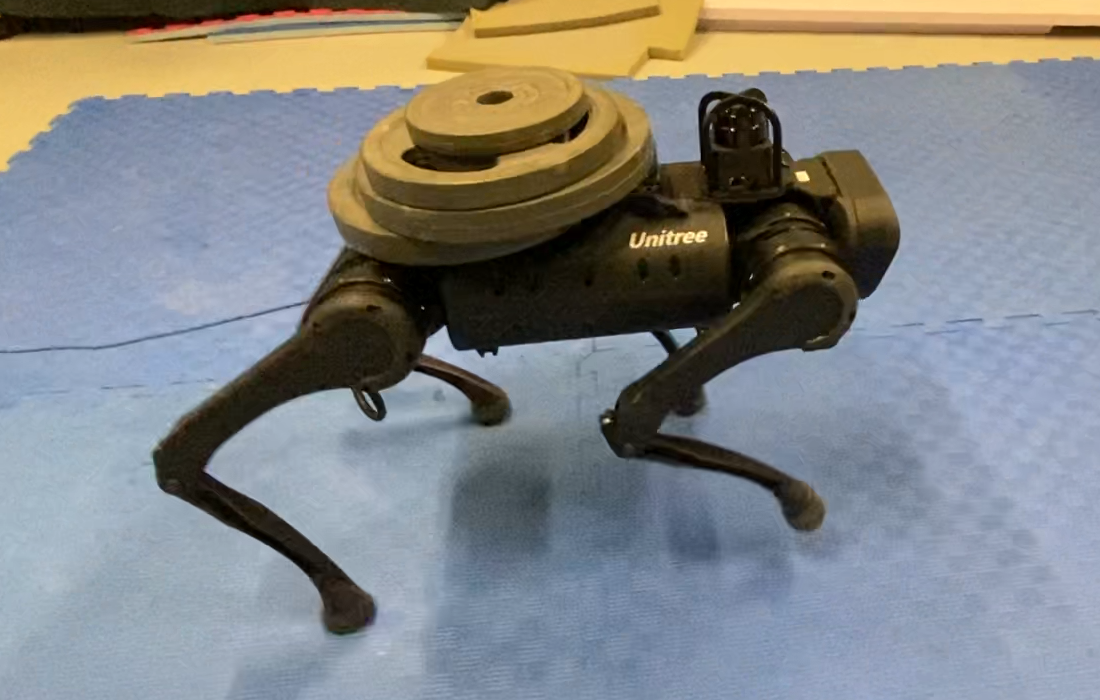} \\
      \vspace{.008in}
      \includegraphics[width=0.25\linewidth]{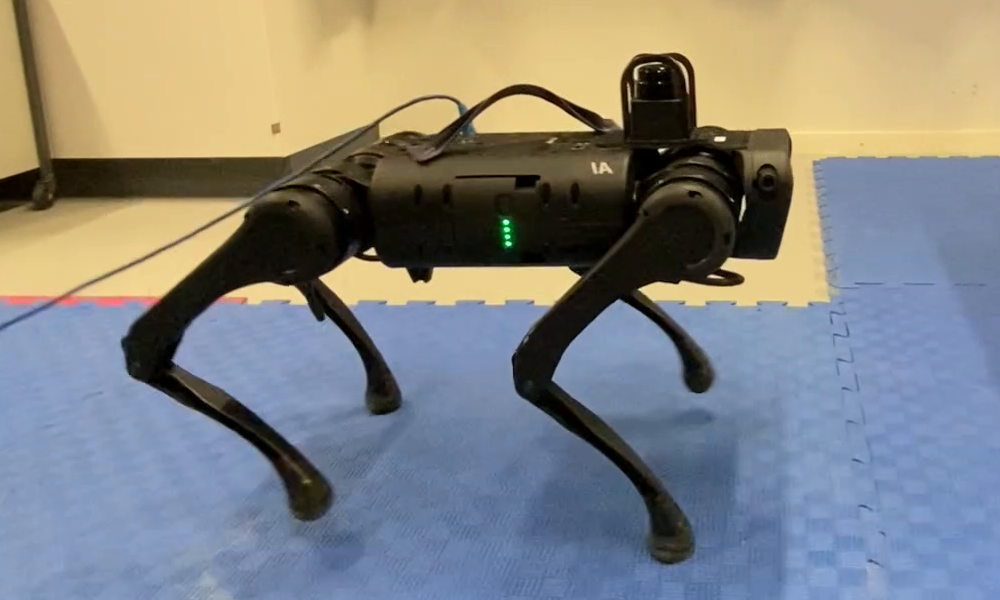}\includegraphics[width=0.25\linewidth]{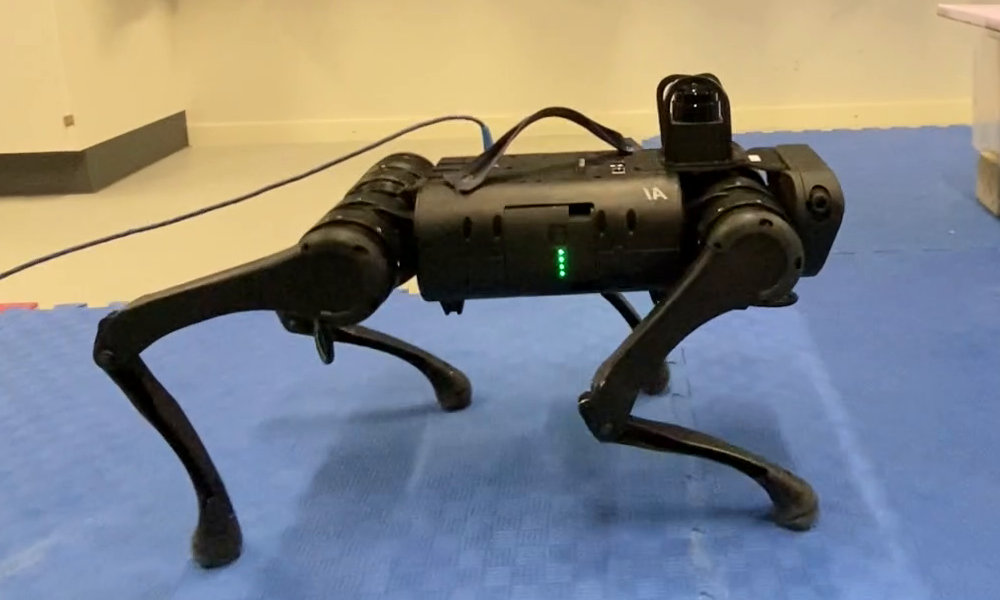}\includegraphics[width=0.25\linewidth]{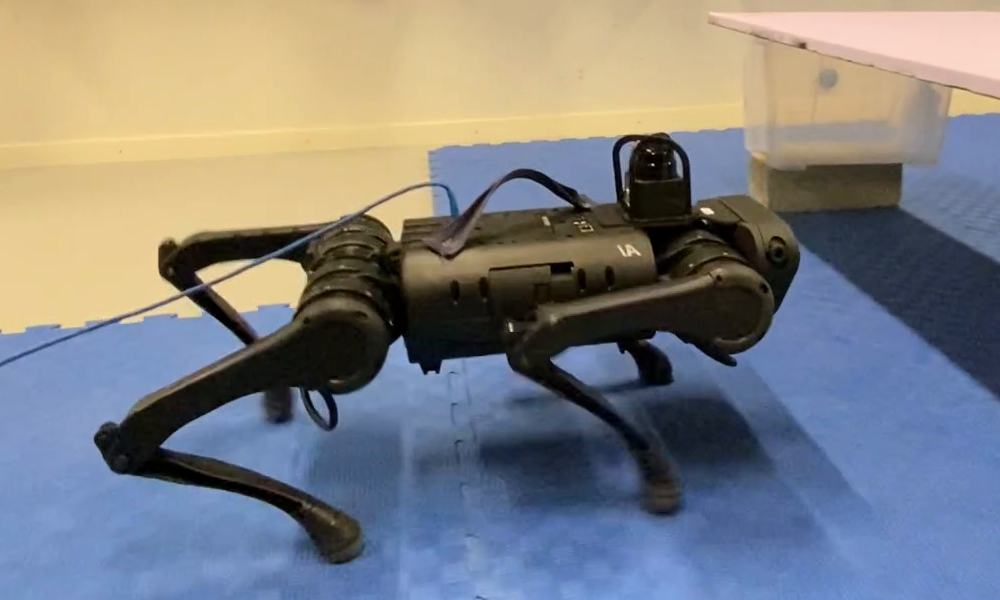}\includegraphics[width=0.25\linewidth]{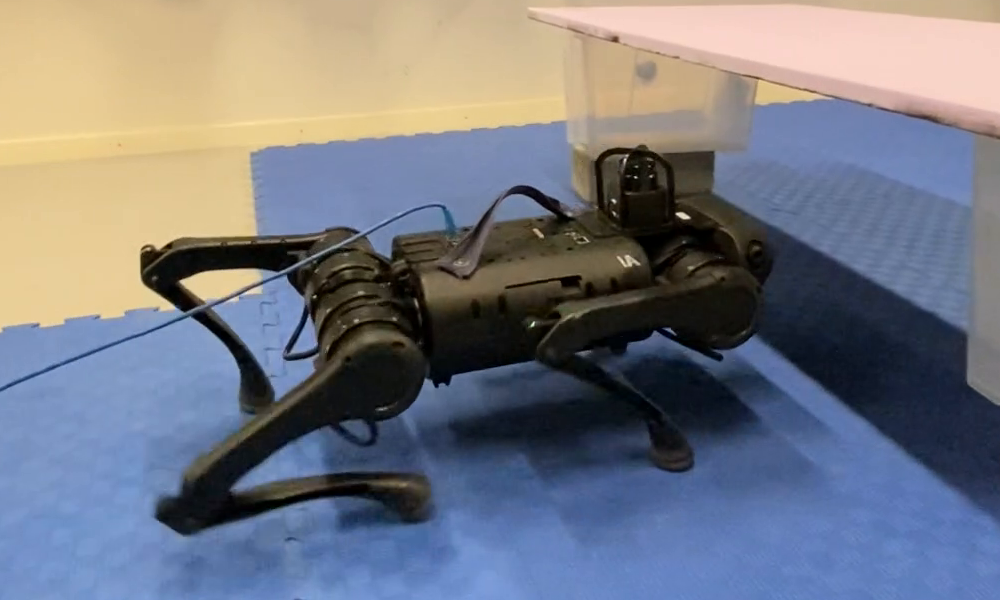} \\
      \vspace{.008in}
      \includegraphics[width=\linewidth]{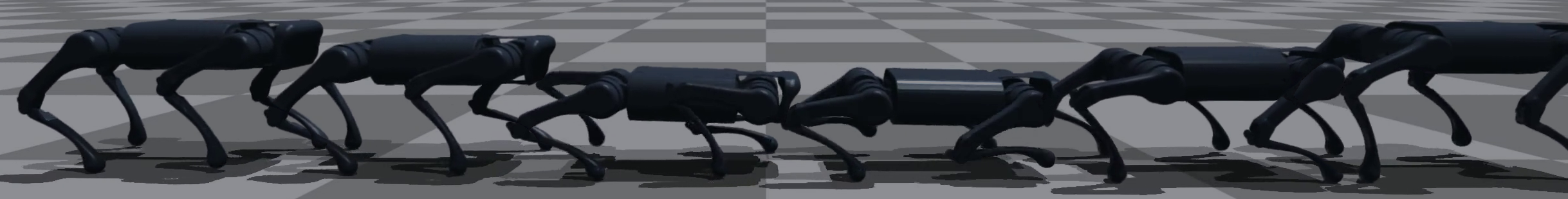}\\
    \vspace{-0.5em}
      \caption{Quadruped locomotion with CPG-RL. \textbf{Top:} crossing uneven terrain (left), and walking at 0.8 \texttt{m/s} with a 13.75 \texttt{kg} load representing 115\% of the nominal quadruped mass (right). 
      \textbf{Bottom:} user-specified body height modulation to crawl underneath a ledge, both experiment and simulation. 
      }
      \label{fig:intro}
      \vspace{-2.0em}
\end{figure}

%--------------------------------------------- Model-Based Control
\subsubsection{Model-Based Control} 
Conventional control approaches have shown impressive performance for real world robot locomotion \cite{dicarlo2018mpc,kim2019highly,sombolestan2021adaptive,bellicoso2018dynamic}. Such methods typically rely on solving online optimization problems (MPC) using simplified dynamics models, which require significant domain knowledge, and may not generalize to new environments not explicitly considered during development (i.e. uneven slippery terrain). There is also considerable bias in the resulting motions due to the (for example) pre-determined foot swing trajectories, and use of Raibert heuristics \cite{raibert1986legged}, which are based on a spring-loaded inverted pendulum model, for foot placement planning. 

%--------------------------------------------- Learning-Based Control
\subsubsection{Learning-Based Control} 
Deep reinforcement learning (DRL) has emerged as an attractive solution to train control policies that are robust to external disturbances and unseen environments in sim-to-real. Similarly to model-based control, recent works have shown successful results in training ``blind'' control policies, with only proprioceptive sensing being mapped to joint commands. For quadrupedal robots, such approaches have resulted in successful locomotion controllers on both flat~\cite{tan2018minitaur,hwangbo2019anymal} and rough terrain~\cite{kumar2021rma,lee2020anymal}. Other works have shown the possibility of directly imitating animal motions~\cite{peng2020laikagoimitation}, and the emergence of different gaits through minimizing energy consumption with DRL~\cite{fu2022energy}. 

To improve on the reactive controllers trained only with proprioceptive sensing as input, recent works are integrating vision into the deep reinforcement learning framework. This has allowed for obstacle avoidance~\cite{yang2022learning} as well as more dynamic crossing of rough terrain through the use of sampling from height maps~\cite{miki2022learning,rudin2022anymalisaac}. Additional works have shown gap crossing~\cite{yu2022visual}, also with full flight phases learned from pixels and leveraging MPC~\cite{margolis2022pixels}. 

%--------------------------------------------- Action Space and Phase in DRL
\subsubsection{Action Space and Phase in DRL} 
Directly mapping from proprioceptive sensing to joint space commands remains a challenging problem, even for deep networks and millions of training samples. In addition to the complexity of specifying desired motions like foot swing height, there are difficulties with the sim-to-real transfer arising from unmodeled dynamics such as the actuators. 
To better structure the learning problem and avoid undesired local optima, recent works are exploring different action spaces (for example directly learning torques~\cite{chen2022learning}, or desired task space positions~\cite{bellegardaIROS19TaskSpaceRL,bellegarda2020robust,bellegarda2021robust}), and in particular encoding a time component (phase) into the framework, such as Policies Modulating Trajectory Generators (PMTG) for Minitaur~\cite{iscen2018policies}, or for ANYmal~\cite{lee2020anymal,miki2022learning}. Additionally, incorporating phase encodings facilitates learning different gaits~\cite{shao2022gait}, and can also be used together with MPC~\cite{yang2022fast}. 

Learning with more biologically inspired approaches with explicit use of CPGs has also been shown in simulation for bipeds~\cite{li2013,kasaei2021cpg}. Li et al.~use reinforcement learning to directly learn the feedback terms of a CPG for a biped to tackle different slopes~\cite{li2013}. Kasaei et al.~use DRL to update the parameters of a CPG-ZMP walk engine and output joint target residuals to adapt to commands and different terrains~\cite{kasaei2021cpg}. For quadrupeds, Shi et al.~learn both the parameters of a foot trajectory generator as well as joint target residuals to locomote in a variety of environments including stairs and slopes~\cite{shi2021reinforcement}.

%--------------------------------------------- Contribution
\subsection{Contribution}
In contrast to previous work, in this paper we propose to use deep reinforcement learning to directly learn the time-varying oscillator intrinsic amplitude and frequency for each oscillator which together forms the Central Pattern Generator. We implement the CPG network with one oscillator per limb, but without explicit couplings between oscillators, similarly to the work of Owaki et al~\cite{owaki2017quadruped}. Couplings between oscillators are known to exist in biological CPGs, but recent work has shown that they might be weaker than previously thought~\cite{owaki2017quadruped,thandiackal2021emergence}, and that sensory feedback and descending modulation might play an important role in inter-oscillator synchronization.
This biology-inspired approach to locomotion additionally mitigates several drawbacks usually encountered each with CPGs and deep reinforcement learning. 

On the CPG side, parameter tuning remains difficult, often necessitating an expert doing so by hand, or the use of genetic algorithms. The tuning of these parameters usually results in only one gait, whereas animals exhibit a continuum of motions at different speeds and directions. Specified (strong) couplings also result in rigid and not completely natural gaits, and the role of sensory feedback and reflexes must then also be tuned and added on top of the CPG. 

On the other hand, when applying deep reinforcement learning algorithms for control tasks, mapping from sensory information to joint commands often results in non-intuitive motions, with great care being needed to properly tune reward functions (i.e., how to specify a desired swing foot height). Additionally, the sim-to-real transfer presents added difficulties from non-modeled dynamics and possible overfitting of simulator physics engine parameters. 

In this work we address all of the above difficulties, and train and deploy our CPG-RL controller on the Unitree A1 quadruped, shown in Figure~\ref{fig:intro}. Some highlights of our method include:

\begin{itemize}
    \item fast training and ease of sim-to-real transfer (i.e. without any domain randomization or added noise in simulation)
    \item a minimal amount of sensory information needed in the observation space (i.e. feedback only from foot contact booleans)
    \item on the fly parameter selection to achieve locomotion at user-specified body heights and foot swing heights, without any needed specification in the DRL framework
    \item robustness to disturbances not seen in training, i.e. uneven terrain, and a dynamically added 13.75 \texttt{kg} load representing 115\% of the nominal quadruped mass
\end{itemize}

The rest of this paper is organized as follows. Section~\ref{sec:background} provides background details on deep reinforcement learning and Central Pattern Generators. Section~\ref{sec:method} describes our design choices and integration of Central Pattern Generators into the deep reinforcement learning framework (CPG-RL). Section~\ref{sec:result} shows results and analysis from learning our controller and sim-to-sim and sim-to-real transfers for several scenarios including minimal observation space, online modulating body height and swing foot ground clearance, and significant disturbances in terrain and added load. Finally, a brief conclusion is given in Section~\ref{sec:conclusion}.

%%%%%%%%%%%%%%%%%%%%%%%%%%%%%%%%%%%%%%%%%%%%%%%%%%%%%%%%%%%%%%%%%%%%%%%%%%%%%%%%
% Background
%%%%%%%%%%%%%%%%%%%%%%%%%%%%%%%%%%%%%%%%%%%%%%%%%%%%%%%%%%%%%%%%%%%%%%%%%%%%%%%%
\section{Background}
\label{sec:background}

%--------------------------------------------- RL
\subsection{Reinforcement Learning}
In the reinforcement learning framework~\cite{sutton1998rl}, an agent interacts with an environment modeled as a Markov Decision Process (MDP). An MDP is given by a 4-tuple $(\mathcal{S,A,P,R})$, where $\mathcal{S}$ is the set of states, $\mathcal{A}$ is the set of actions available to the agent, $\mathcal{P}: \mathcal{S} \times \mathcal{A} \times \mathcal{S} \rightarrow \mathbb{R}$ is the transition function, where $\mathcal{P}(s_{t+1} | s_t, a_t)$ gives the probability of being in state $s_t$, taking action $a_t$, and ending up in state $s_{t+1}$, and  $\mathcal{R}: \mathcal{S} \times \mathcal{A} \times \mathcal{S} \rightarrow \mathbb{R}$ is the reward function, where $\mathcal{R}(s_t,a_t,s_{t+1})$ gives the expected reward for being in state $s_t$, taking action $a_t$, and ending in state $s_{t+1}$. The agent goal is to learn a policy $\pi$ that maximizes its expected return for a given task.

%--------------------------------------------- CPGs
\subsection{Central Pattern Generators}
\label{sec:cpg_intro}
While a number of neural oscillators have been proposed to implement CPG circuits, here we focus on the amplitude-controlled phase oscillators from~\cite{ijspeert2007salamander} used to produce both swimming and walking behaviors on a salamander robot: 
\begin{align}
\ddot{r}_i &= a \left(\frac{a}{4} \left(\mu_i - r_i \right) - \dot{r}_i \right) \label{eq:salamander_r} \\
\dot{\theta}_i &= \omega_i +\sum_{j}^{} r_j w_{ij} \sin(\theta_j - \theta_i - \phi_{ij}) \label{eq:salamander_theta}
\end{align}
where $r_i$ is the current amplitude of the oscillator, $\theta_i$ is the phase of the oscillator, $\mu_i$ and $\omega_i$ are the intrinsic amplitude and frequency, $a$ is a positive constant representing the convergence factor. Couplings between oscillators are defined by the weights $w_{ij}$ and phase biases $\phi_{ij}$.

Regardless of the particular oscillator selection, several choices exist for the number of oscillators in the network, as well as how to map the output back to joint commands. For example, one oscillator can be used for each joint to directly produce motion in joint space~\cite{ijspeert2007salamander,sprowitz2013cheetah}, or the oscillator can be formed in task space and mapped to joint commands with inverse kinematics~\cite{righetti08}. Thus, this formulation results in numerous design decisions and parameters that must be tuned, commonly by hand or through evolutionary algorithms such as CMA-ES~\cite{hansen2016cma}. Notably, this tuning generally results in specific fixed gaits which may not be robust to external disturbances.

%%%%%%%%%%%%%%%%%%%%%%%%%%%%%%%%%%%%%%%%%%%%%%%%%%%%%%%%%%%%%%%%%%%%%%%%%%%%%%%%
% Method 
%%%%%%%%%%%%%%%%%%%%%%%%%%%%%%%%%%%%%%%%%%%%%%%%%%%%%%%%%%%%%%%%%%%%%%%%%%%%%%%%
\section{Learning Central Pattern Generators}
\label{sec:method}

\begin{figure*}[!t]
      \centering
    \includegraphics[width=0.77\linewidth]{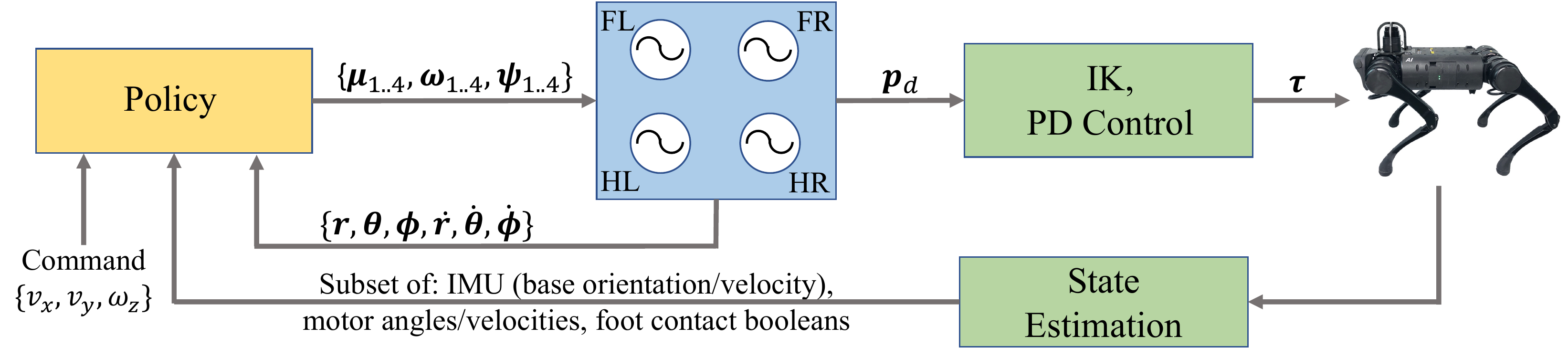}
    \vspace{-0.7em}
      \caption{The CPG-RL control architecture. Given an observation consisting of velocity commands, a subset of proprioceptive measurements, and the current CPG states, the policy network selects CPG parameters $\mu$, $\omega$, and $\psi$ for each leg $i$ (Front Left (FL), Front Right (FR), Hind Left (HL), Hind Right (HR)). The CPG states are mapped to desired foot positions $\bm{p}_d$, which are then converted to desired joint angles with inverse kinematics, and finally tracked with joint PD control to produce torques $\bm{\tau}$. The control policy selects actions at 100 Hz, and all other blocks operate at 1 kHz.}
      \label{fig:control_diagram}
      \vspace{-1.7em}
\end{figure*}

In this section we describe our CPG-integrated reinforcement learning framework and design decisions for learning locomotion controllers for quadruped robots. At a high level, the agent learns to modulate the CPG parameters for each leg with deep reinforcement learning. This allows for adaptation based on sensory feedback along with the knowledge of the current CPG state. This type of learning represents an approximation of motor learning in animals, namely how higher brain centers such as the motor cortex and the cerebellum learn to send modulation signals to CPG circuits in the spinal cord. The control diagram for our method is shown in Figure~\ref{fig:control_diagram}, and we discuss all components below.  

%-------------------------------------Action Space
\subsection{Action Space} 
\label{sec:action_space}

\begin{figure}[!tpb]
      \centering
      \includegraphics[width=0.85\linewidth]{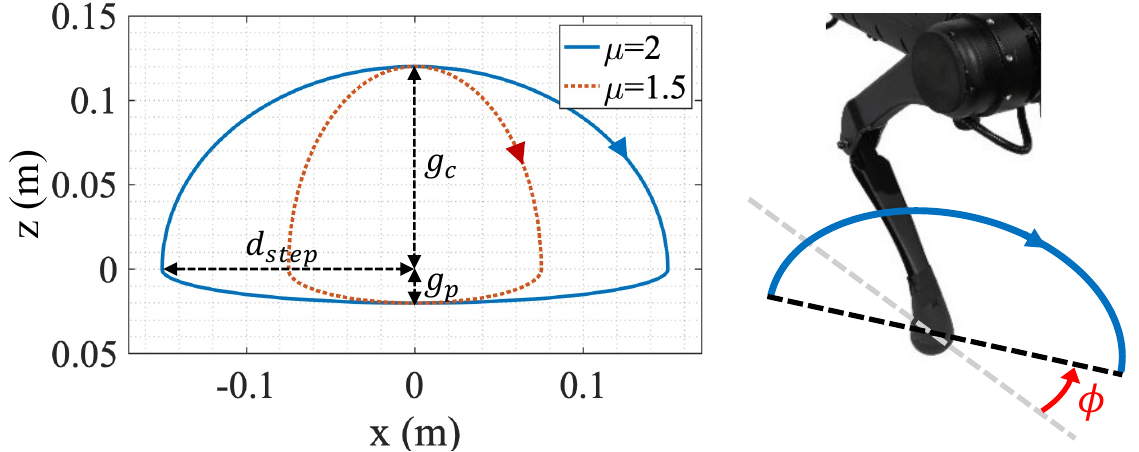} \\
    \vspace{-0.6em}
      \caption{Mapping the CPG states to Cartesian foot positions. At left in the XZ-plane: ground clearance ($g_c$), ground penetration ($g_p$), max step length ($d_{step}$) are design parameters, whereas CPG states $r$ and $\theta$ control amplitude and frequency. Two trajectories are shown representing the mapping for two converged amplitude set points, $\mu$, which the agent can directly modulate to vary the step length. At right, CPG state $\phi$ controls the orientation of the trajectory with respect to the body, shown for the Front Right (FR) leg.}
      \label{fig:cpg_traj}
      \vspace{-2.2em}
\end{figure}

We first consider the coupled oscillators in Equations~\ref{eq:salamander_r} and~\ref{eq:salamander_theta} (one for each leg, or $i \in \{1,2,3,4\}$), whose output we wish to map to foot trajectories in Cartesian space similar to~\cite{righetti08}. Such a strategy poses two issues: 1) the coupling severely biases the rhythm into potentially unnatural motions, and 2) with only two variables the motions are restricted to a single plane. Inspired by animals, which frequently deviate and transition between various gaits and produce omnidirectional motions, we solve both issues by removing any explicit coupling ($w_{ij}=0$), with the intuition that the agent should learn to coordinate its limbs on its own (as opposed to being fixed by the CPG coupling topology), and we add another state variable, $\phi$, to orient motion in any direction (see right of Figure~\ref{fig:cpg_traj}). Thus, for each limb we define the following oscillator: 
\begin{align}
\ddot{r}_i &= a \left(\frac{a}{4} \left(\mu_i - r_i \right) - \dot{r}_i \right) \label{eq:rl_r}\\
\dot{\theta}_i &= \omega_i  \label{eq:rl_theta}\\
\dot{\phi}_i &= \psi_i  \label{eq:rl_phi}
\end{align}

In contrast to recent works making use of phase in reinforcement learning, we propose an action space to directly modulate the intrinsic oscillator amplitude and phases, by learning to modulate $\mu_i$, $\omega_i$, and $\psi_i$ for each leg. This allows the agent to adapt each of these states online in real-time (for instance increasing the amplitude to step farther or accelerating a leg movement during swing), compared with the more traditional CPG approach of optimizing for only a single set of fixed parameters. Additionally, we hypothesize that this framework will lead to more interpretable results, as we can directly observe how the agent modulates the oscillators depending on feedback from both the environment as well as from the current oscillator states (for example in contrast to directly learning joint commands). Thus, for the omnidirectional locomotion task, our action space can be summarized as $\vec{a} = [\bm{\mu}, \bm{\omega}, \bm{\psi}] \in \mathbb{R}^{12}$. The agent selects these parameters at 100 Hz, and we use the following limits for each input during training: $\mu \in [1, 2]$, $\omega \in [0, 4.5]$ Hz, and $\psi \in [-1.5, 1.5]$ Hz, and $a=150$.  

To map from the oscillator states to joint commands, we first compute corresponding desired foot positions, and then calculate the desired joint positions with inverse kinematics. The desired foot position coordinates are given as follows:
\begin{align}
x_{i,\text{foot}} &= -d_{step} (r_i-1) \cos(\theta_i) \cos(\phi_i) \\
y_{i,\text{foot}} &= -d_{step} (r_i-1) \cos(\theta_i) \sin(\phi_i) \\
z_{i,\text{foot}} &= \begin{cases}
    -h + g_c\sin(\theta_i) & \text{if } \sin(\theta_i) > 0 \\
    -h + g_p\sin(\theta_i) & \text{otherwise}
\end{cases}
\end{align}
\noindent where $d_{step}$ is the maximum step length, $h$ is the robot height, $g_c$ is the max ground clearance during swing, and $g_p$ is the max ground penetration during stance. A sample visualization of the foot trajectory for a set of these parameters is shown at left of Figure~\ref{fig:cpg_traj}. 
These parameters make it possible to specify behaviors that are in general difficult to learn when directly outputting joint commands. For example, specifying a foot swing height would usually necessitate keeping track of a history of states and exasperates the temporal credit assignment problem of reinforcement learning. With our framework, we randomly sample $h$, $g_c$, and $g_p$ during training (i.e. the agent has no explicit observation of these parameters) to learn continuous behavior, which then allows the user to specify both a robot height and swing foot ground clearance during deployment. 

%-------------------------------------Observation Space
\subsection{Observation Space}
\label{sec:obs_space}

We consider three observation spaces in this work with varying amounts of proprioceptive sensing: full ($obs_{full}$), medium ($obs_{med}$) and minimal ($obs_{min}$). Our purpose is to investigate how much the locomotor performance depends on the richness of the observation space, and also to identify which information is sufficient to reach reasonable performance. This investigation is also interesting for neuroscience to explore which type of sensor modality is more important than others for generating stable gaits. The CPG states are always in the observation space, and compared with the proprioceptive sensing which is subject to measurement noise from onboard sensors, the CPG states are always known. This provides a source of stability to the method and we believe eases the sim-to-real transfer. 

\noindent \textbf{Full observation:} The full observation consists of velocity commands and measurements reasonably available with proprioceptive sensing, and are becoming standard in DRL approaches. These include the body state (orientation, linear and angular velocities), joint state (positions, velocities), and foot contact booleans. The last action chosen by the policy network and CPG states $\{\bm{r,\dot{r},\theta,\dot{\theta}, (\phi, \dot{\phi})}\}$ are concatenated to the proprioceptive measurements. 

\noindent \textbf{Medium observation:} The medium observation removes the joint state and last action from the full observation. This observation space is chosen to show that joint information is actually not necessary for omnidirectional locomotion through our method. Other states remain the same (i.e.~velocity commands, body state, foot contact booleans, and CPG states). 

\noindent \textbf{Minimal observation:} The minimal observation space consists only of foot contact booleans and the CPG states $\{\bm{r,\dot{r},\theta,\dot{\theta}}\}$. This observation space shows that coordination between limbs can be accomplished with very little sensing at all, with the only environmental feedback being from foot contact booleans. The idea for this space is inspired from the force feedback term in traditional CPGs shown to coordinate transitions between gaits~\cite{owaki2013simple,owaki2017quadruped}, also known as \textit{Tegotae}-based control~\cite{owaki2017minimal}. The importance of contacts and limb loading has also been shown by Ekeberg and Pearson in a simulation of cat locomotion~\cite{ekeberg2005computer}. For this observation space, the task is only to move forward at a particular desired velocity $v_{b,x}^{*}$. 

%------------------------------------- Reward
\subsection{Reward Function} 
\label{sec:reward_function}

We design our reward function to track desired body velocity commands in the body frame $x$ and $ y$ directions as well as a desired yaw rate $\omega_{b,z}^{*}$. We include terms to minimize other undesired body velocities as well as penalize the work (aiming at keeping the body stable and minimizing the energy consumption). More precisely, the reward function is a summation of the following terms:

\begin{itemize}
    \item linear velocity tracking, body $x$ direction:  $f(v_{b,x}^{*} - v_{b,x})$
    \item linear velocity tracking, body $y$ direction:  $f(v_{b,y}^{*} - v_{b,y})$
    \item angular velocity tracking (body yaw rate):  $f(\omega_{b,z}^{*} - \omega_{b,z})$
    \item linear velocity penalty in body $z$ direction: $-v_{b,z}^2$
    \item angular velocity penalty (body roll and pitch rates): $-||\bm{\omega}_{b,xy}||^2$
    \item work: $-|\bm{\tau} \cdot (\dot{\bm{q}}^t-\dot{\bm{q}}^{t-1})|$
\end{itemize}

\noindent where $(\cdot)^{*}$ represents a desired command, and $f(x) := \exp{(-\frac{||x||^2}{0.25}) } $. These terms are weighted with $0.75 dt$, $0.75 dt$, $0.5dt$, $2dt$, $0.05dt$, $0.001dt$, where $dt=0.01$ is the control policy time step. Notably, as discussed in Section~\ref{sec:action_space}, we do not need to put any additional terms on foot swing time. Compared with other learning-based approaches for omnidirectional locomotion, this is a simple set of terms to properly specify desired behavior. 

\begin{table}[tpb]
\centering
\footnotesize
\caption{PPO Hyperparameters and neural network size.}
\vspace{-0.6em}
\resizebox{0.71\linewidth}{!}{%
\begin{tabular}{ c c  }
Parameter & Value \\
\hline
Batch size & 98304 (4096x24) \\
Mini-bach size & 24576 (4096x6)\\
Number of epochs & 5\\
Clip range & 0.2\\
Entropy coefficient & 0.01\\
Discount factor & 0.99\\
GAE discount factor & 0.95\\
Desired KL-divergence $kl^*$ & 0.01\\
Learning rate $\alpha$ & adaptive\\
Number of hidden layers (all networks) & 3 \\
Number of hidden units per layer & [512, 256, 128] \\
Activation & elu \\
\hline
\end{tabular} }\\
\label{table:ppo}
\vspace{-2.5em}
\end{table}

%-------------------------------------Training Details
\subsection{Training Details}
\label{sec:training_details}

We use Isaac Gym~\cite{isaacgym,rudin2022anymalisaac} with PhysX as our physics engine and training environment, and the Unitree A1 quadruped~\cite{unitreeA1}. This framework allows for high throughput, where we simulate 4096 A1s in parallel on a single NVIDIA RTX 3070 GPU. 
We use PPO~\cite{ppo} to train the policy, and the relevant hyperparameters and neural network architecture details (multilayer perceptron with 3 hidden layers) are listed in Table~\ref{table:ppo}. With this framework, similar to in~\cite{rudin2022anymalisaac}, we can learn control policies within minutes. 

The maximum episode length is 20 seconds, and the environment resets for an agent if the base or a thigh comes in contact with the ground. The terrain is always flat during training. With each reset, we sample new parameters $h$ and $g_c$ for the mapping from oscillator states to joint commands so the agent can learn to locomote at varying body heights and step heights. New velocity commands $\{v_{b,x}^{*},v_{b,y}^{*},\omega_{b,z}^{*}\}$ are sampled every 5 seconds. Although we find that domain randomization is not strictly needed to perform a sim-to-real transfer, unless specified we randomize the following parameters during training (kept constant during an episode):
\begin{itemize}
    \item \textbf{ground coefficient of friction} varied in [0.3, 1]
    \item \textbf{limb mass} varied within 20\% of nominal values
    \item \textbf{added base mass} up to 5 \texttt{kg}
    \item \textbf{external push} of up to 0.5 \texttt{m/s} applied in a random direction to the base every 15 seconds
\end{itemize}
No noise is added to the observation. 

The control frequency of the policy is 100 Hz, and the torques computed from the desired joint positions are updated at 1 kHz. The equations for each of the oscillators (Eqns.~\ref{eq:rl_r}-\ref{eq:rl_phi}) are thus also integrated at 1 kHz. The joint PD controller gains are $K_p=100,\ K_d=2$. For the sim-to-real transfer, all proprioceptive information is measured from the Unitree sensors. 

%%%%%%%%%%%%%%%%%%%%%%%%%%%%%%%%%%%%%%%%%%%%%%%%%%%%%%%%%%%%%%%%%%%%%%%%%%%%
% Results
%%%%%%%%%%%%%%%%%%%%%%%%%%%%%%%%%%%%%%%%%%%%%%%%%%%%%%%%%%%%%%%%%%%%%%%%%%%%
\section{Experimental Results and Discussion}
\label{sec:result}

In this section we report results from learning locomotion controllers with CPG-RL. We seek to evaluate the necessity of sensory information as defined by the three observation spaces, the difficulty of the sim-to-real transfer, the interpretability of the resulting policy, and the robustness to various disturbances not seen during training. Snapshots of some of our results are shown in Figure~\ref{fig:intro}, and the supplementary video shows clear visualizations of the discussed experiments. 

\input{tex/exp_results}
\input{tex/sim2sim}

%%%%%%%%%%%%%%%%%%%%%%%%%%%%%%%%%%%%%%%%%%%%%%%%%%%%%%%%%%%%%%%%%%%%%%%%%%%%%%%%
% Conclusion
%%%%%%%%%%%%%%%%%%%%%%%%%%%%%%%%%%%%%%%%%%%%%%%%%%%%%%%%%%%%%%%%%%%%%%%%%%%%%%%%
\section{Conclusion}
\label{sec:conclusion}
In this work we have presented CPG-RL, a framework for learning and modulating intrinsic oscillator amplitudes and frequencies to coordinate rhythmic behavior among limbs to achieve quadruped locomotion. Our results have shown that this method results in fast training and ease of sim-to-real transfer, where we show successful transfers with no domain randomization and only minimal sensory feedback. Additionally, the framework allows the user to easily specify (on the fly and/or in training) desired legged robot quantities like body height and swing foot ground clearance, the latter of which can be challenging to specify through reward shaping in end-to-end learning frameworks. The framework also proved robust to disturbances not seen in training, for example A1 was able to walk over uneven terrain or with added mass 2.75x greater than in training (115\% of nominal robot mass) in hardware, and 250\% of the nominal robot mass in sim-to-sim. To the best of our knowledge, this represents the highest robustness against loads so far achieved on the Unitree A1 quadruped. 

In terms of neuroscience, the use of DRL will allow us in the future to explore questions that are still open in animal motor control, namely the exact roles and interactions of descending pathways, interoscillator couplings within CPG networks, and sensory feedback in gait generation. The results presented here suggest (i) that descending pathways are more effective at modulating locomotion by acting on the CPG circuits rather than directly on muscles (CPG-RL performs better than $Joint$ $PD$), (ii) that stable locomotion can be obtained with non-existent (or weak) interoscillator couplings, and (iii) that sensing contact (or loading) in the limb is one of the most important sensory information. This last point is in agreement with the conclusion of a neuromechanical simulation of cat locomotion~\cite{ekeberg2005computer}. 

\section*{Acknowledgements}
We would like to thank Milad Shafiee and Alessandro Crespi for assisting with hardware setup and experiments. 

%%%%%%%%%%%%%%%%%%%%%%%%%%%%%%%%%%%%%%%%%%%%%%%%%%%%%%%%%%%%%%%%%%%%%%%%%%%%%%%%
\bibliographystyle{IEEEtran}
\bibliography{refs}

\end{document}

%% file: tex/exp_results.tex
%--------------------------------------------------------- Sim-to-Real Experimental Results
\subsection{Sim-to-Real Experimental Results}

%--------------------------------------------------------- CPG state modulation
\subsubsection{CPG State Modulation}
\label{sec:result_cpg_state}

In the video, we examine how the agent trained with CPG-RL coordinates and modulates the CPG states to produce locomotion. We verify how similar the resulting gait is compared with fixed open-loop CPG gaits (trot, walk, pace) as generated by Equations~\ref{eq:salamander_r} and~\ref{eq:salamander_theta}, tuned for locomotion at particular frequencies.

\begin{figure}[!t]
    \centering
    \includegraphics[width=0.7\linewidth]{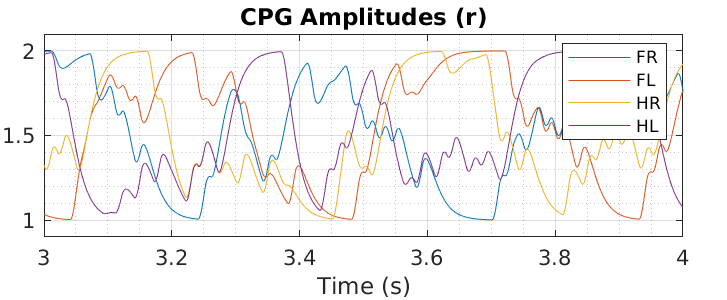} \\
    \includegraphics[width=0.7\linewidth]{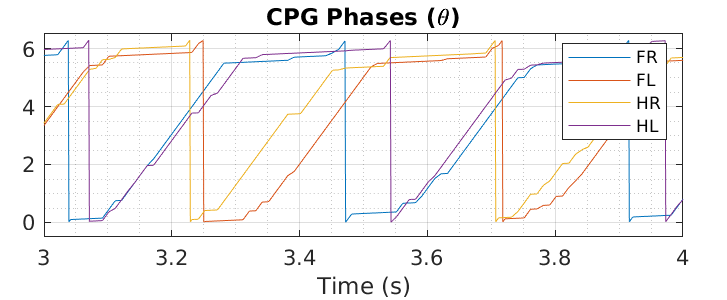} \\
    \vspace{-0.5em}
    \caption{CPG states for 1 second of a deployed policy locomoting at 0.8 \texttt{m/s}. An approximate trot gait can be observed, with faster swing time ($0 \leq \theta \leq \pi $) than stance time. Coordination between the amplitude and phase variables can be seen, noticeably increasing amplitude in stance phase to push backward and propel the quadruped forward.}
    \label{fig:cpg_r_theta}
    \vspace{-1.2em}
\end{figure}

\begin{figure}[!t]
    \centering
    \includegraphics[width=\imgHalf\linewidth]{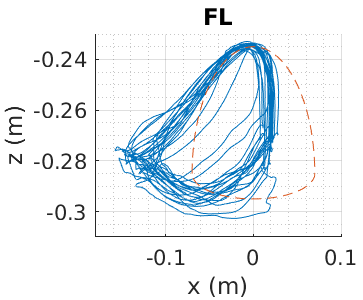}\hspace{\imgHalfSpace\linewidth}\includegraphics[width=\imgHalf\linewidth]{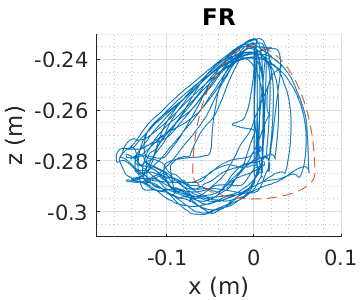} \\
    \includegraphics[width=\imgHalf\linewidth]{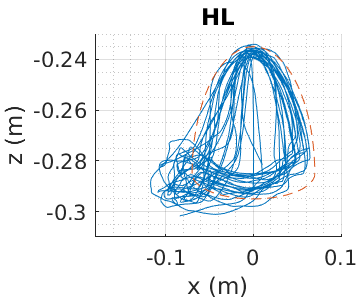}\hspace{\imgHalfSpace\linewidth}\includegraphics[width=\imgHalf\linewidth]{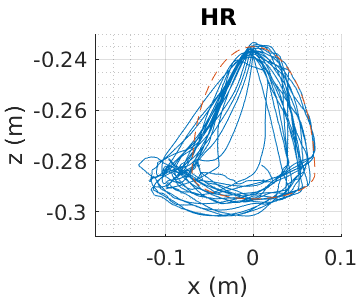} \\
    \vspace{-0.5em}
    \caption{Leg frame foot XZ trajectories for a deployed policy locomoting at 0.8 \texttt{m/s} with dynamically added mass. Significant modulations can be seen to the sample task space trajectory represented by the dotted line. Notably, the amplitude for the front feet shifts the trajectory to lie mostly behind the hip. 
    }
    \label{fig:cpg_xz}
    \vspace{-2.3em}
\end{figure}

We command the agent to walk forward at 0.8 \texttt{m/s} while adding variable mass up to 13.75 \texttt{kg}. One second of the CPG amplitude and phase plots is shown in Figure~\ref{fig:cpg_r_theta}, where we observe an approximate trot gait with a cycle of approximately 0.5 seconds. The swing duration (when $0 \leq \theta \leq \pi $) can be observed to be lower than the stance duration, as is typical of quadruped animals. It is additionally apparent that there is coordination between phase and amplitude, the latter of which is not quite as periodic or constant, implying the agent uses it more to adapt to sensory feedback. In general however the amplitude appears higher when in stance phase, showing the agent uses this time to push backwards and propel the quadruped forward. This result is also apparent by examining the leg frame foot $XZ$ trajectories, shown in Figure~\ref{fig:cpg_xz}. Notably, the trajectories for both front feet are significantly modulated from the nominal task space trajectory used in the open-loop trot gait in the video, where now the foot is primarily underneath and behind the hip in the leg frame. Compared with the fixed open-loop CPG gaits, CPG-RL produces a continuum of more natural gaits that are less rigid, more efficient, more robust, and have lower frequency.

%--------------------------------------------------------- Min Obs space
\subsubsection{Observation Space Effects}
\label{sec:result_obs}

As can be seen in Figure~\ref{fig:intro} and the video, we achieve robust sim-to-real transfer with and without joint state information (i.e.~using either the full or medium observation spaces, $obs_{full}$ and $obs_{med}$). This result holds for omnidirectional motions at varying velocities for scenarios including uneven terrain, push recovery, and significant added loads encompassing 115\% of the nominal robot mass (see Section~\ref{sec:result_mass}). This result is attributable to the presence of the oscillator states in the observation space, and its mapping back to the task space trajectories. 

We are also able to transfer the CPG-RL $obs_{min}$ policy, for which the only feedback from the environment that the agent observes is through the contact booleans, and is trained without any simulator noise or randomization.  While the agent has no direct observation of its body velocity, it can relate the frequencies of the neural oscillator states (which it has direct control over) to the reward it receives at each time step, to track 0.5 \texttt{m/s} in the body $x$ direction while keeping all other velocities 0. This result supports that coordination between limbs is possible through very little sensory feedback \cite{owaki2013simple,owaki2017quadruped,owaki2017minimal}. 

We also tested training policies without the foot contact feedback, i.e. with only CPG states in the observation space, but the agent was unable to learn to locomote at any fixed velocity. This shows that some sort of feedback is necessary to coordinate locomotion.

%--------------------------------------------------------- body height / swing height 
\subsubsection{Body Height and Swing  Foot Height Online Modulation}
As discussed in Section~\ref{sec:action_space}, our framework naturally allows the user to specify the body height and foot ground clearance through setting parameters $h$ and $g_c$ for the mapping from oscillator states to desired foot positions. The agent has no explicit knowledge of these parameters, but in the full observation space it can find a relationship from observing the direct effects on the joint positions. 

Once training is complete a user can change either of these parameters in real time, as the agent continues to locomote at its velocity commands. In Figure~\ref{fig:intro} and in the video we show an example of lowering the body height $h$ from its nominal height of 0.30 \texttt{m} to 0.19 \texttt{m} in order to crawl underneath a ledge, and then increasing it back to 0.30 \texttt{m} on the other side. We also test varying the foot ground clearance $g_c$ online, changing it from 0.03 \texttt{m} to 0.12 \texttt{m} (almost half of the nominal robot height), and then back down to 0.03 \texttt{m}. 

%--------------------------------------------------------- uneven terrain / added mass 
\subsubsection{Uneven Terrain} 
\label{sec:result_terrain}
As discussed in Section~\ref{sec:training_details}, we train our policies only on flat terrain, though the coefficient of friction is varied. Notably, the rigid body dynamics used in simulation are not exactly representative of the hardware, as the A1 feet undergo significant deformations on contact, in addition to other sim-to-real considerations such as the lack of motor modeling. We tested adding debris such as soft foam and hard styrofoam (see Figure~\ref{fig:intro} and video) and found the policy to be robust to such terrain. These materials are very light and easy to kick and crumple by A1, and even when the material got stuck or rolled up, causing swing feet to catch and impeding progress, it did not immediately fall. 

\subsubsection{Added Mass}
\label{sec:result_mass}
We also tested adding varying mass to the robot while it was trotting at various velocities (0.3-0.8 \texttt{m/s}). In all cases there was no noticeable drop in performance, which can be attributed to both the robustness of the method as well as the high joint gains we are able to use. We achieved trotting at 0.8 \texttt{m/s} with 13.75 \texttt{kg} (115\% of nominal robot mass) dynamically added to the robot, which was all of the available mass we had in the lab. To the best of our knowledge, this represents the highest robustness against loads achieved on A1. Other comparisons include RMA~\cite{kumar2021rma} achieving up to a 12 \texttt{kg} load at lower velocity, a model-based method achieving 11 \texttt{kg} while standing~\cite{sombolestan2021adaptive}, and the Unitree default model-based controller has a max rating of 5 \texttt{kg}~\cite{unitreeA1}. 

Remarkably, we used the domain randomization explained in Section~\ref{sec:training_details}, which only added up to 5 \texttt{kg} in simulation, again showing significant robustness to out of distribution disturbances. Additionally, as we achieve the same results  with both the medium and full observation spaces, we show that joint state information is actually not necessary for dynamic locomotion with our method. 

%% file: tex/sim2sim.tex
\subsection{Sim-to-Sim Comparison with Learning in Joint Space}
\label{sec:result_comparison}

To evaluate the benefits of our approach, we compare CPG-RL with a standard joint space training pipeline, which takes as input proprioceptive measurements, and outputs joint space offsets from a nominal resting position using the same method from~\cite{rudin2022anymalisaac}, which we call $Joint$ $PD$. The observation space contains all proprioceptive sensing used in the full observation space by CPG-RL, without the oscillator states. All training hyperparameters, environment details, and neural network architectures remain the same. To train the $Joint$ $PD$ baseline, we compare using both (a) the same reward function as described in Section~\ref{sec:reward_function}, as well as (b) the more complex \textit{special} reward function from~\cite{rudin2022anymalisaac}, which additionally includes terms for orientation, joint acceleration, joint velocity, joint torque, action rate, collisions, feet air time, and base height. 

The video shows training curves for learning locomotion policies with CPG-RL as well as with the $Joint$ $PD$ baseline trained with both reward functions (a) and (b). While all three methods provide similar returns as training progresses, the policies trained with CPG-RL produce the most natural looking gaits, are easiest to interpret, and allow the user to set swing foot ground clearance. In contrast, $Joint$ $PD$ policies trained with the same reward function (a) result in unnatural gaits that overfit the simulator dynamics (shown in the video), and are unlikely to transfer well in sim-to-real. $Joint$ $PD$ policies trained with the more complex \textit{special} reward function (b) can achieve more natural gaits, however tracking accuracy is lower (Tables~\ref{table:vx_comp} and~\ref{table:omni_comp_v2}), and it is not possible to directly control the swing foot ground clearance. 

\input{tex/sim2sim_tables}

%---------------------------------------------------------- 
Quantitatively, we evaluate and compare the performance of CPG-RL as well as $Joint$ $PD$  policies~\cite{rudin2022anymalisaac} in a sim-to-sim transfer from Isaac Gym with the PhysX physics engine to Gazebo with the ODE physics engine. While a successful sim-to-sim transfer does not necessarily guarantee a successful sim-to-real transfer, sim-to-sim allows safely testing many policies without risking damaging the hardware, as used in recent work to verify the agent has not overfit the training simulator's dynamics before sim-to-real transfers~\cite{bellegarda2021robust,chen2022learning}. While there is no noise added to the observation space in Isaac Gym which uses ground truth data, in Gazebo we simulate the state estimation (i.e. Kalman Filter) as used on the real hardware. For each method (CPG-RL or $Joint$ $PD$~\cite{rudin2022anymalisaac}) and observation space ($obs_{full}$, $obs_{med}$, $obs_{min}$), we also compare performance when training with and without randomization as described in Section~\ref{sec:training_details}.

%--------------------------------------------------------- 
\subsubsection{Sim-to-sim Forward Locomotion}  
We first train policies (at least 10) for each method and observation space with the goal of tracking only forward velocities in the body $x$ direction, namely $v_{b,x}^{*} \in [0.2, 1]$ (\texttt{m/s}). Table~\ref{table:vx_comp} shows sim-to-sim tracking performance of desired command  $v_{b,x}^{*} = 0.5$ (\texttt{m/s}), where we present the mean velocity, duty factor, gait period, as well as body height and foot ground clearance. The best tracking performance is for CPG-RL with both full and medium observation spaces. The mean duty factor and gait period are much higher for policies trained with CPG-RL, suggesting greater stability. Compared with $Joint$ $PD$ policies which learn to locomote at a fixed body height and ground clearance, all CPG-RL policies can vary height and foot ground clearance online.

%--------------------------------------------------------- 
\subsubsection{Sim-to-sim with Added Loads}
We take the same policies and repeat the transfers while adding loads to the robot and still commanding $v_{b,x}^{*} = 0.5$ (\texttt{m/s}).  Figure~\ref{fig:mass_comp} shows the mean velocity tracking performance for added masses from 0 to 30 \texttt{kg}, in increments of 3 \texttt{kg}. All policies, whether trained with noise (i.e. up to 5 \texttt{kg} in Isaac Gym) or without, are able to make at least some forward progress with loads up to 9 \texttt{kg}. The points labeled with $*$s indicate that performance is not guaranteed: the robot either stops or falls down, with increasing probability for higher loads. The dashed lines show the performance of policies trained without any noise in Isaac Gym, which notably still allows all policies trained with CPG-RL to locomote with 15 \texttt{kg}, even for the minimal observation space $obs_{min}$. Interestingly, the policies trained with CPG-RL and $obs_{med}$ perform better than policies trained with $obs_{full}$. Under higher disturbances, the joint states are an additional source of noise and may take value combinations unseen in training, which may shift the expected distribution the agent has learned to map between observations and actions. The results show that CPG-RL allows sim-to-sim transfer with loads representing 250\% of the nominal robot mass, while trained with noise of only up to 42\% of the nominal robot mass.

%--------------------------------------------------------- 
\subsubsection{Sim-to-sim Omnidirectional Commands}
We next train policies (at least 10) for each method and observation space to track omnidirectional commands in the following ranges: $v_{b,x}^{*} \in [-1, 1]$ (\texttt{m/s}), $v_{b,y}^{*} \in [-1, 1]$ (\texttt{m/s}), $\omega_{b,z}^{*} \in [-1, 1]$ (\texttt{rad/s}). 
Table~\ref{table:omni_comp_v2} shows sim-to-sim tracking performance of various commanded omnidirectional velocities within these ranges. The data shows that CPG-RL policies can closely track desired forward, lateral, and angular velocities, as well as combinations of these, even when trained without noise. In contrast, we note that in addition to not tracking the commands as accurately, transferring the omnidirectional $Joint$ $PD$ policies comes with several added difficulties compared to CPG-RL.  
As the training curve converges, there is a very small window (can be fewer than 50 iterations) for which the $Joint$ $PD$ policies are able to transfer sim-to-sim, which is also not consistent across different random seeds.
If training continues, while the average return does not increase, the resulting policies become more and more unnatural as the agent exploits the simulator dynamics (we observe tiny steps with the front limbs, and both rear limbs in the air), and are unable to transfer sim-to-sim. We also observe that training $Joint$ $PD$ policies requires much more tuning and design decisions, including reward function tuning, dynamics randomization parameter tuning, possible motor modeling, etc. which overall results in longer training times compared with CPG-RL. 

%--------------------------------------------------------- 
\subsubsection{Joint PD Observation Spaces}
While possible to learn omnidirectional locomotion with $Joint$ $PD$ and $obs_{med}$, such policies are unable to transfer due to learning high frequency small-step gaits that exploit the simulator dynamics, lacking feedback from the joint states. Training in joint space with $obs_{min}$ is unable to learn any locomotion policy.

%% file: tex/sim2sim_tables.tex
\renewcommand{\tabcolsep}{2pt}

\begin{table}[t!]
\centering
\footnotesize
\caption{Sim-to-sim tracking of velocity command $v_{b,x}^{*} = 0.5$ \texttt{m/s} with different observation spaces and methods trained to locomote for $v_{b,x}^{*} \in [0.2, 1]$ (\texttt{m/s}) (except for CPG-RL with $obs_{min}$, which is trained to exclusively track $v_{b,x}^{*} = 0.5$ \texttt{m/s}). We compare performance when training with/without randomization (Sec.~\ref{sec:training_details}).
}
\vspace{-0.8em}
\resizebox{0.85\linewidth}{!}{
\begin{tabular}{ | c | c | c | c | c | c | c || c | c | }
\hline
Mean Quantity & \multicolumn{6}{c||}{CPG-RL} & \multicolumn{2}{c|}{Joint PD [25]} \\
\hline
Obs. Space & \multicolumn{2}{c|}{$obs_{full}$}  & \multicolumn{2}{c|}{$obs_{med}$}   & \multicolumn{2}{c||}{$obs_{min}$}  &  \multicolumn{2}{c|}{$obs_{full}$}   \\
\hline
\begin{tabular}{@{}c@{}} Training \\ Randomization? \end{tabular} & X & \checkmark & X & \checkmark & X & \checkmark & X & \checkmark \\
\hline
$v_{b,x}$ (\texttt{m/s}) & 0.494 & 0.493 & 0.491 & 0.488 & 0.460 & 0.472 & 0.486 & 0.471 \\
\hline
Duty Factor & 1.74 & 2.10 & 1.48 & 1.75 & 2.63 & 2.83 & 0.87 & 0.89 \\
\hline
Period (\texttt{s}) & 0.75 &  0.70 & 0.70 & 0.75 & 0.90 & 0.75 & 0.55 & 0.60\\
\hline
\begin{tabular}{@{}c@{}} Body \\ Height (\texttt{m}) \end{tabular} & \multicolumn{6}{c||}{0.17 - 0.30} & 0.29 & 0.30 \\
\hline
\begin{tabular}{@{}c@{}} Foot Ground\\ Clearance (\texttt{m}) \end{tabular} & \multicolumn{6}{c||}{0.03 - 0.12} & 0.025 & 0.02 \\
\hline
\end{tabular}}
\label{table:vx_comp}
\vspace{-1.2em}
\end{table}

\begin{table}[t!]
\centering
\footnotesize
\caption{Sim-to-sim tracking ability of omnidirectional commands with different observation spaces and methods trained to locomote in the following ranges: $v_{b,x}^{*} \in [-1, 1]$ (\texttt{m/s}), $v_{b,y}^{*} \in [-1, 1]$ (\texttt{m/s}), $\omega_{b,z}^{*} \in [-1, 1]$ (\texttt{rad/s}). We compare performance when training with/without randomization (Sec.~\ref{sec:training_details}).}
\vspace{-0.8em}
\resizebox{0.78\linewidth}{!}{
\begin{tabular}{ | c | c | c | c | c || c | c | }
\hline 
Command : Actual & \multicolumn{4}{c||}{CPG-RL} & \multicolumn{2}{c|}{Joint PD [25]} \\
\hline
Obs. Space & \multicolumn{2}{c|}{$obs_{full}$}  & \multicolumn{2}{c||}{$obs_{med}$}   &  \multicolumn{2}{c|}{$obs_{full}$}   \\
\hline
\begin{tabular}{@{}c@{}} Training \\ Randomization? \end{tabular} & X & \checkmark & X & \checkmark & X & \checkmark \\
\hline
$v_{b,x}^{*}$ = 0.5 (\texttt{m/s}) & 0.447 & \textbf{0.488} & 0.545 & 0.466 & 0.467 & 0.421 \\
\hline
$v_{b,y}^{*}$ = 0.5 (\texttt{m/s}) & 0.483 & 0.481 & 0.359 & \textbf{0.510} & 0.260 & 0.483 \\
\hline
$v_{b,x}^{*}$ = 0.5 (\texttt{m/s}) & 0.470 & \textbf{0.492} &  0.437 & 0.446 & 0.435 & 0.418 \\
$v_{b,y}^{*}$ = 0.5 (\texttt{m/s}) & 0.476 & \textbf{0.490} & 0.436 & 0.478 & 0.366 & 0.459 \\
\hline
$\omega_{b,z}^{*}$ = 0.6 (\texttt{rad/s}) & 0.463 & \textbf{0.578} & 0.544 & 0.537 & 0.418 & 0.537 \\
\hline
\end{tabular}} \\
\label{table:omni_comp_v2}
\vspace{-1.2em}
\end{table}

\begin{figure}[!t]
\centering
\includegraphics[width=\linewidth]{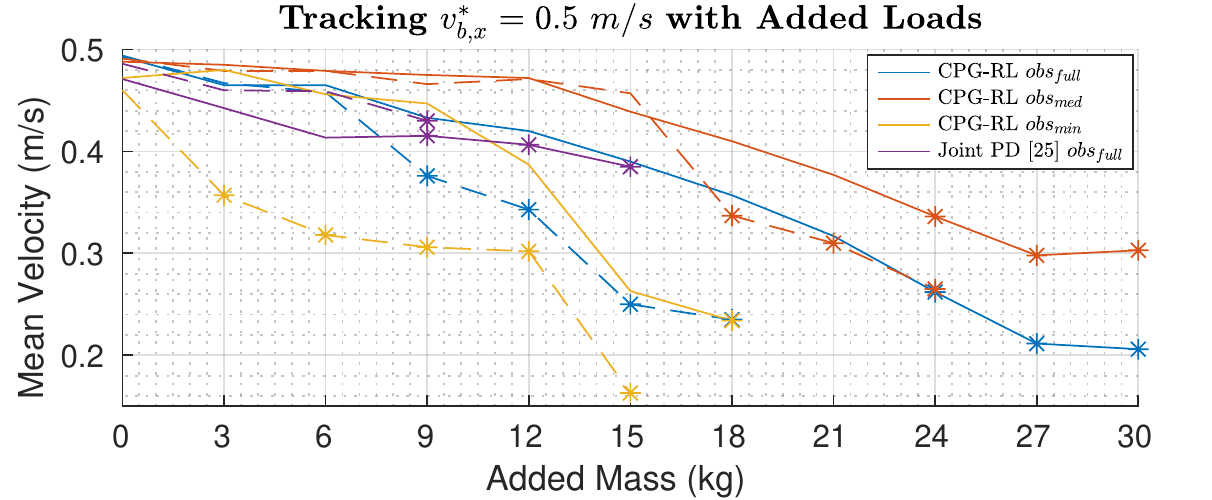} \\
\vspace{-0.7em}
      \caption{Sim-to-sim mean velocity for command $v_{b,x}^{*} = 0.5$ \texttt{m/s} under increasing added loads. The dashed lines represent the same method/observation trained without any noise in Isaac Gym.  The $*$s represent when performance is not guaranteed: the robot sometimes falls down, or cannot make consistent forward progress.}
      \label{fig:mass_comp}
      \vspace{-2.2em}
\end{figure}